\documentclass[sigconf]{acmart}

\renewcommand\footnotetextcopyrightpermission[1]{} 
\usepackage{algorithm, algorithmic, multirow}

\def\BibTeX{{\rm B\kern-.05em{\sc i\kern-.025em b}\kern-.08emT\kern-.1667em\lower.7ex\hbox{E}\kern-.125emX}}
    
\begin{document}

%
\title[Anomaly Detection in Sequences of Attributed Graph Edges]{ADSAGE: Anomaly Detection in Sequences of Attributed Graph Edges applied to insider threat detection at fine-grained level}

%
\author{Mathieu Garchery}
\email{mathieu.garchery@atos.net}
\affiliation{
	\institution{Atos Information Technology}
	\streetaddress{Otto-Hahn-Ring 6}
	\city{Munich}
	\country{Germany}
	\postcode{81739}
}
\affiliation{
	\institution{University of Passau}
	\streetaddress{Innstr. 41}
	\city{Passau}
	\country{Germany}
	\postcode{94032}
}

\author{Michael Granitzer}
\email{michael.granitzer@uni-passau.de}
\affiliation{
  \institution{University of Passau}
  \streetaddress{Innstr. 41}
  \city{Passau}
  \country{Germany}
  \postcode{94032}
}

%

%
\begin{abstract}
	Previous works on the CERT insider threat detection case have neglected graph and text features despite their relevance to describe user behavior. Additionally, existing systems heavily rely on feature engineering and audit data aggregation to detect malicious activities. This is time consuming, requires expert knowledge and prevents tracing back alerts to precise user actions. To address these issues we introduce ADSAGE to detect anomalies in audit log events modeled as graph edges. Our general method is the first to perform anomaly detection at edge level while supporting both edge sequences and attributes, which can be numeric, categorical or even text. We describe how ADSAGE can be used for fine-grained, event level insider threat detection in different audit logs from the CERT use case. Remarking that there is no standard benchmark for the CERT problem, we use a previously proposed evaluation setting based on realistic recall-based metrics. We evaluate ADSAGE on authentication, email traffic and web browsing logs from the CERT insider threat datasets, as well as on real-world authentication events. ADSAGE is effective to detect anomalies in authentications, modeled as user to computer interactions, and in email communications. Simple baselines give surprisingly strong results as well. We also report performance split by malicious scenarios present in the CERT datasets: interestingly, several detectors are complementary and could be combined to improve detection. Overall, our results show that graph features are informative to characterize malicious insider activities, and that detection at fine-grained level is possible.
\end{abstract}

%
%


%

%
\maketitle

\section{Introduction}

A recent whitepaper \cite{insider-threat-report-2018} reveals that 90\% of organizations feel vulnerable to insider threats, i.e. legitimate users who abuse their access rights to IT systems to conduct malicious activities such as data theft, sabotage and misuse. Worse, according to the same source, 53\% of organizations confirmed having been targeted in the last 12 months. Insider threats are particularly harmful to organizations as the attacker usually possesses knowledge about his environment, which could help evade detection and increase attack impact.

To address this issue and foster research on insider threats, in 2013 the CERT (Computer Emergency Response Team) of CMU's Software Engineering Institute has released a corpus of synthetic datasets \cite{Glasser2013, CERT_dataset}. In a field where public datasets are extremely scarce due to confidentiality reasons, this release has triggered a large amount of academic publications. However important research gaps remain; we focus on three of them.

First, most existing detection systems do not integrate all audit data sources provided in the CERT datasets. Especially heterogeneous features like graph and text are often discarded as they are not supported by many anomaly detection methods, unlike numeric and categorical features. This is surprising as works addressing other insider threat settings suggest that graph and text data can be quite helpful. For example, graph features can be used to model communication between members of an organization \cite{Eberle2010, Okolica2008} or accesses to resources \cite{Senator2013, Chen2012}, in which anomalous patterns can reveal malicious insiders. Through sentiment analysis and psychometric measures, text data can help detecting risk factors such as worker discontentment \cite{Kandias2013, Brown2013}.

Second, existing insider threat detection systems heavily rely on data aggregation and feature engineering \cite{Tuor2017, Gavai2015, Bose2017}. Indeed, this strategy can be effective, nevertheless at the cost of alert traceability. For instance in \cite{Tuor2017}, users are assigned an anomaly score for all their daily activities, thus determining specifically which action(s) lead to an alert is not straightforward.

Third, despite all resources concerning the CERT insider threat use case being public, unfortunately no standard benchmark methodology has emerged. Existing works use different metrics and data subsets for evaluation, rendering performance comparison difficult. 

We address the CERT insider threat use case while tackling these three issues. Concerning the first -- support of graph and text data -- we introduce ADSAGE for anomaly-based intrusion detection supporting numeric, categorical, but also graph and text attributes. In particular, we leverage graph features by modeling user events (equivalent to log lines) as graph edges representing interactions between entities. For instance, an email being sent corresponds to an edge from the sender to the receiver. Edges can be augmented with attributes to provide context, such as the time the email was sent or its text content. Using a recurrent neural network (RNN), ADSAGE is able to take into account sequences of events. A feed-forward neural network (FFNN) is used simultaneously to predict the validity of events and output anomaly scores accordingly. Given such attributed graph edges, we show how to use ADSAGE to uncover insider threats in the CERT datasets. Note that ADSAGE's applicability is not limited to insider threat detection: our method can be used for anomaly detection in sequences of attributed graph edges in general. To the best of our knowledge, no existing method for anomaly detection at edge level supports both edge sequences and attributed edges.

Regarding the second issue (alert traceability), our method operates at event (i.e. log line) level with a unique data source. This allows flagging anomalies at a fine-grained level and reduces the need for feature engineering and data aggregation. However, it is important to note that direct performance comparison with existing systems like \cite{Tuor2017}, which leverage multiple audit data sources, is unfair. This study's primary goal is rather to determine whether detection at event level without aggregation is feasible at all. Using the different malicious insider scenarios in the CERT setting as our threat model, we empirically determine which data sources are relevant to detect different threat scenarios.

Concerning the third issue (absence of standard evaluation methodology), as an effort towards benchmark standardization we adopt the evaluation setting from \cite{Tuor2017}, chosen for its business-realistic metrics. We complement our results with an evaluation on real authentications from the LANL cybersecurity datasets \cite{LANL-paper, LANL-dataset}.

Our contributions can be summarized as follows:
\begin{itemize}
	\item We introduce ADSAGE, a general method to detect anomalies in sequences of graph edges with numeric, categorical and text attributes.
	\item We show how to apply ADSAGE for fine-grained event level detection in the CERT insider threat use case, enhancing alert traceability and reducing the need for feature engineering.
	\item We empirically evaluate our method on three log data sources from the CERT datasets (logon, email and web) and on LANL authentication events. Our experiments suggest that ADSAGE is effective for logon and email data. By reporting detection results for individual threat scenarios, we show which audit data sources are relevant to detect each scenario.
\end{itemize}

\section{Problem setting and approach}
\subsection{CERT insider threat use case}\label{CERT_dataset}
The CERT insider threat datasets \cite{Glasser2013, CERT_dataset} contain synthetic data representing the activity of users within a large organization. Available audit data sources include logon events, email traffic, web browsing traces, file access logs, usage of removable devices as well as LDAP information describing the organization hierarchy and user roles. We focus on events that are straightforwardly represented as interactions between two entities, i.e. graph edges: logon (user to computer), email (sender to receivers) and web browsing (user to web domain) events. In our evaluation, we use version 6.2 of the CERT dataset, which contains one example of each scenario. Note that this represents an extremely unbalanced problem with an anomaly rate in the order of $10^{-6}$ at event level or $10^{-5}$ at user-day level (i.e. when aggregating all data sources daily for each user).

Our threat model consists of insider threat scenarios which are described as follows in the CERT documentation \cite{CERT_dataset}:
\begin{enumerate}
	\item User who did not previously use removable drives or work after
	hours begins logging in after hours, using a removable drive, and
	uploading data to wikileaks.org. Leaves the organization shortly
	thereafter.
	\item User begins surfing job websites and soliciting employment from a
	competitor. Before leaving the company, they use a thumb drive (at
	markedly higher rates than their previous activity) to steal data.
	\item System administrator becomes disgruntled. Downloads a keylogger and
	uses a thumb drive to transfer it to his supervisor's machine. The
	next day, he uses the collected keylogs to log in as his supervisor
	and send out an alarming mass email, causing panic in the
	organization. He leaves the organization immediately.
	\item A user logs into another user's machine and searches for
	interesting files, emailing to their home email. This behavior occurs
	more and more frequently over a 3 month period.
	\item A member of a group decimated by layoffs uploads documents to
	Dropbox, planning to use them for personal gain.
\end{enumerate}

\subsection{Anomaly detection at event level} 

We address the CERT insider threat use case through an anomaly detection perspective, i.e. we aim at modeling normal user behavior to detect deviations from this norm. Such anomalies are then considered as insider threat alarms. While this perspective has been widely adopted for intrusion detection, unlike existing systems our approach is to perform detection at fine-grained event level. In the CERT insider threat use case, one event (i.e. log line) represents an elementary user action and usually contains features to describe its context. For instance, an event can represent a logon to a particular computer and features can be the event time or the device used. Our goal is to assign an anomaly score to each audit event.

The primary reason to perform detection at fine-grained event level is to enhance alert traceability. Intrusion detection systems are typically not used as standalone solution, but rather perform a first selection of suspicious activities to be further scrutinized by security analysts. In this context, flagging anomalies at fine-grained event level eases traceability, as analysts will be able to determine exactly which user action lead to an alert. On the contrary, using a system like \cite{Tuor2017}, an anomaly score is assigned to a whole day of user activity, thus when an alarm is raised the question of which exact elements triggered it remains open. A second advantage is that data aggregation and feature engineering efforts are greatly reduced compared to systems like \cite{Tuor2017, Gavai2015, Bose2017}. As we will show next, except for time features (which we transform only to reflect their periodical nature), our methods use audit event attributes without further preprocessing.

\section{Methods}

\subsection{Seq2one baseline}
\label{seq2one-baseline}

\begin{figure}
	\centering
	\includegraphics[width=\columnwidth]{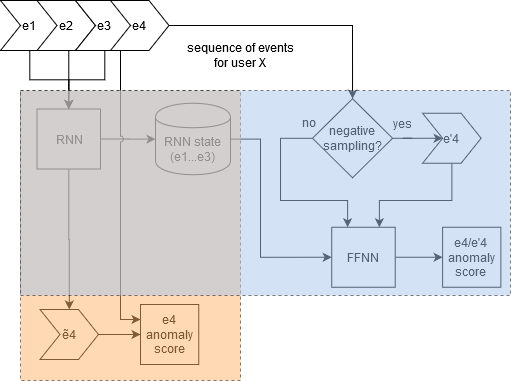}
	\caption{Joint training architecture for the recurrent (RNN) and feed-forward neural network (FFNN) used in ADSAGE. This example shows the anomaly score prediction for event $e4$ given previous events $e1$...$e3$. Note that ADSAGE components are delimited by the blue area, while the seq2one baseline corresponds to the orange area. Seq2one is trained on normal events only, predicts the entire next event $\tilde{e4}$ and compares it to the real $e4$ to derive an anomaly score. ADSAGE is trained on both normal and anomalous events being generated with negative sampling (e.g. $e\sp{\prime}4$).}
	\Description{Training workflow showing how seq2one and ADSAGE are trained and predict anomaly scores.}
	\label{fig-rnn-ffnn-joint-model}
\end{figure}

To detect insider threats at event level, we adapt DeepLog \cite{Du2017}, a log line anomaly detector for system traces. As we take into account only one audit data source at a time, we only keep DeepLog's event features prediction module. It computes an anomaly score based on the error between predicted and observed value for the next event. We adapt the error function to support numeric and categorical attributes. For numeric features, mean squared error is used and for categorical attributes the error is $1-p$ where $p$ is the probability of the true category, obtained by applying the softmax function. Each error is then normalized by using its quantile (e.g. 0.99 if the error is greater than 99\% of observed errors for this feature).  Quantiles are finally averaged to obtain an event anomaly score. This method is referred to as "seq2one" and corresponds to the orange area in figure \ref{fig-rnn-ffnn-joint-model}.

Unfortunately, our previous experiments on the CERT datasets have shown that seq2one gives poor threat recall. One plausible explanation is that user behavior in far less predictable than machine behavior, hence predicting the next event is much more difficult with user activity traces than with system logs used in \cite{Du2017}. This motivates us to extend DeepLog to better learn the distinction between normal and anomalous behavior by introducing ADSAGE.

\subsection{ADSAGE: Anomaly Detection in Sequences of Attributed Graph Edges}
As predicting the exact features of the next event is difficult for user generated events, we propose ADSAGE, a method focusing on predicting the validity of graph edges. In the following, we detail how events can be represented as attributed graph edges (section \ref{adsage-representing-events}) and how ADSAGE is trained to predict the validity of such events (section \ref{adsage-training}), by relying on negative sampling (section \ref{adsage-neg-sampling}).

\subsubsection{Representing events as attributed graph edges}\label{adsage-representing-events} 
\begin{figure*}
	\centering
	\includegraphics[width=\textwidth]{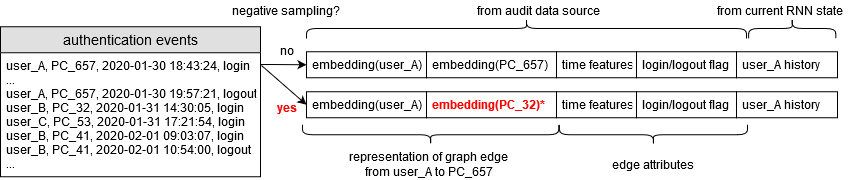}
	\caption{Representation of audit events as attributed graph edges in ADSAGE on the example of authentication events. An authentication event corresponds to a user-computer edge and is represented as the concatenation of user and computer embeddings. Edge attributes can be added as well (i.e. time features, flag indicating logon or logoff). The user history (RNN state) representing previous activities of current user A is also appended. Here, the first authentication event is encoded to be processed by ADSAGE's FFNN. On the right, the upper row represents the normal event. The lower row represents an anomalous version of the same event, obtained with negative sampling on the edge destination (indicated by the asterisk). The accessed computer in the anomalous event (PC\_32) is selected at random among all computers never used by user A.}
	\Description{Illustration of how a log line is represented as attributes graph edge in ADSAGE. The figure shows how different attributes of an authentication record are concatenated to obtain a vector representation.}
	\label{fig-event-to-edge}
\end{figure*}

In ADSAGE events are represented as attributed graph edges. Figure \ref{fig-event-to-edge} shows an example on authentication events similar to logs from CERT. An authentication event is an interaction between a user and a computer, corresponding to an edge in the graph of users and computers. In ADSAGE, this edge is represented as  the concatenation of its source (user) and destination (computer) entities. As ADSAGE is based on neural network models, an embedding layer is used for each entity feature. Thus source and destination embeddings are optimized according to the prediction task (described in section \ref{adsage-training}).

In addition to its source and destination entity, an edge can also have features extracted from its event context (e.g. time features and logon/logoff attributes in figure \ref{fig-event-to-edge}). Different types of features are possible: numeric values, categorical attributes (as one-hot or embedding representation) or even text content (via pre-trained word embeddings).

Note that ADSAGE can be easily extended to the case where an event has multiple sources and/or destinations. Events with a fixed and limited number of sources or destinations can be represented as concatenation of corresponding embeddings. For events with a high and/or varying number of sources or destinations, it is possible to use an embedding bag layer \cite{embedding-bag} (i.e. an embedding layer with pooling function such as average or max) to obtain a fixed-length representation.

\subsubsection{Training FFNN and RNN jointly to learn edge validity}\label{adsage-training}
To perform anomaly detection in sequences of attributed edges, we use a combination of sequence-to-one RNN (similarly to seq2one) and feedforward neural network (FFNN), both trained jointly. Given a sequence of events for a given user, the RNN is trained to predict the next event and outputs an RNN state representing the event history up to this instant. The RNN uses a mixture of mean squared error (for numeric features), cross-entropy (for one-hot encoded features) and cosine loss (for embeddings). 

The RNN state encoding history of previous events is used as input for the FFNN, together with the next event. The FFNN is trained to predict whether the edge representing the next event is valid, which is formulated as a binary classification task using cross-entropy loss. 

Figure \ref{fig-rnn-ffnn-joint-model} shows the full architecture with RNN and FFNN, and algorithm \ref{algo-adsage} details how both are trained simultaneously. Both our seq2one baseline and ADSAGE maintain a separate RNN state for each user. With this mechanism each user event sequence can be modeled individually while the model is trained on all users. Note that ADSAGE is trained on both normal and anomalous events (generated by negative sampling, see section \ref{adsage-neg-sampling}) and outputs anomaly scores directly while seq2one is trained on observed events only and predicts entire events. As shown later in evaluation, these differences allow ADSAGE to better detect anomalous events.

\begin{algorithm}
	\caption{Training ADSAGE}
	\label{algo-adsage}
	\begin{algorithmic}
		\STATE function train\_ADSAGE (n\_epochs, batches):
		\FOR{ep in n\_epochs}
		\STATE reset\_rnn\_states() // set all user RNN states to zero 
		\FOR{ffnn\_x, ffnn\_y, rnn\_x, rnn\_y in batches}
		\STATE // ffnn\_x: next events including negative samples
		\STATE // ffnn\_y: event validity labels
		\STATE // rnn\_x: sequences of true (positive) events
		\STATE // rnn\_y\_true: next true event for each sequence in rnn\_x
		\STATE // \textbf{FFNN step}
		\STATE // retrieve current RNN states for FFNN inputs
		\STATE ffnn\_states = get\_user\_rnn\_states(ffnn\_x) 
		\STATE // concat input events and user histories
		\STATE ffnn\_input = \{ffnn\_x, ffnn\_states\} 
		\STATE // predict validity of events (anomaly scores)
		\STATE ffnn\_y\_pred = FFNN(ffnn\_input) 
		\STATE backprop(ffnn\_y\_pred, ffnn\_y)
		\STATE // \textbf{RNN step}
		\STATE // retrieve current RNN states for RNN inputs
		\STATE rnn\_states = get\_user\_rnn\_states(rnn\_x)
		\STATE // predict the next event for each sequence
		\STATE rnn\_y\_pred, rnn\_states = RNN(rnn\_x, rnn\_states) 
		\STATE backprop(rnn\_y\_pred, rnn\_y)
		\STATE save\_user\_rnn\_states(rnn\_states) 
		\ENDFOR
		\ENDFOR
	\end{algorithmic}
\end{algorithm}

\subsubsection{Generating anomalous edges through negative sampling}\label{adsage-neg-sampling}
In order to get negative examples for the event validity classification task (i.e. anomalous edges), we artificially replace the destination entity through negative sampling (see figure \ref{fig-event-to-edge}). In a negative event, the destination entity should be anomalous in the sense that interactions from the source entity are usually not observed. In practice we randomly draw a destination entity (e.g. computer for logons) from the set of destinations never accessed from the source entity (e.g. user) during the training period, while other edge attributes are left unchanged. We use a constant negative sampling rate of $0.5$ (i.e. for one positive event, we generate a corresponding negative event), however this value could be tuned as desired.

\section{Evaluation}\label{eval}
We first present our general evaluation methodology in section \ref{eval-methodology}, then we describe results obtained on the CERT insider threat dataset in section \ref{eval-cert} and on authentication events from the LANL cybersecurity dataset in section \ref{eval-lanl}.
\subsection{Evaluation setting}\label{eval-methodology}
\subsubsection{Recall-based metrics}
\label{metrics}
For our evaluation, we use recall-based metrics introduced in \cite{Tuor2017}: recall curves and cumulative recall at budget $k$ ($CR_k$). These metrics are realistic from the perspective of an organization with a fixed budget to investigate alerts generated by an insider threat detection system. The organization's daily budget $k$ represents the number of (most suspicious) users to be investigated each day. If a malicious user is investigated on a given day, all his malicious activities conducted that day are considered as detected. Recall (at budget $k$) $R_k$ is computed as the recall of malicious users per day, averaged over all test days (days with no malicious activity are ignored).

Note that although ADSAGE detects threats at \textit{event} level, recall at budget is computed at \textit{user-day} level, i.e. in terms of number of anomalous users detected for a given day. The first reason to do so is to allow a comparison with \cite{Tuor2017}. The second is that when a user is investigated following an alert, the investigator will have to review the entire user activity (at least the whole user-day) to have sufficient context to come to a decision.

$R_k$ reflects detection performance at a fixed daily investigation budget. To assess performance across multiple budgets, $R_k$ can be plotted against $k$ up to a maximum budget $k_{max}$ to obtain a recall curve. Such curve can be summarized with normalized cumulative recall computed as $CR_k$ = $\sum_{i=0}^{i=k}$ $R_i$ $/n$, where $n$ is the number of budget steps. $CR_k$ can be seen as an approximation of the area under recall curve up to budget $k$.

\subsubsection{Baselines}
In each evaluation setting, we benchmark ADSAGE against 3 different types of baselines. The first is "seq2one" which uses an RNN model to predict the features of next event given previous events (see section \ref{seq2one-baseline}).

Simple rule-based classifiers constitute the second type of baselines. These models are not expected to be competitive for insider threat detection in practice, but they should be outperformed by ADSAGE to ensure that detected anomalies are not trivial. For example, if each user is assigned a computer, a simple rule is to consider all authentication attempts to a different computer as anomalous. Similar rules can be used for other types of events, the general pattern being that an edge from a source to a destination entity is flagged as anomalous if it was not observed in the train set, and as normal otherwise.

The last baseline we use is SedanSpot \cite{Eswaran2018}, a general anomaly detection method applicable to sequences of graph edges. SedanSpot takes into account the timestamp of each edge, but does not support additional edge attributes. Note that SedanSpot and rule-based methods are deterministic and do not depend on initialization. This is why we report exact performance metrics for these methods, unlike for ADSAGE and seq2one.

\subsubsection{Tuning ADSAGE hyperparameters}
For each dataset we optimize ADSAGE's hyperparameters. Most of them are related to the underlying neural networks (RNN and FFNN). We tune following hyperparameters: number of timesteps, hidden units and layers in the RNN, batch size, dimension of embeddings used to represent graph features, learning rate and use different types of pre-trained word embeddings (for datasets containing text features). To speed up training on large datasets, we reduce training set size by sampling users randomly. The user sample rate is another hyperparameter to tune, and one can also choose to sample only from users presenting no malicious behavior. Testing is always performed on all users. We tune one hyperparameter at a time to determine its optimal value, then combine all best parameter values as final configuration. Although this process does not take into account dependencies between hyperparameters, it is much faster than extensive grid search. For each evaluation setting we report the optimal configuration found.

\subsection{Insider threat detection in CERT dataset}
\label{eval-cert}

Using the CERT dataset version 6.2, we perform the same train/test data split as \cite{Tuor2017} to compare our results to theirs. We report cumulative recall (CR, see section \ref{metrics}) at budgets 400 and 1000 and at maximum budget 4000 for completeness. We also report recall metrics based on detecting \textit{all} threats present in the test set, i.e. malicious activity across \textit{all event types}, including log data sources not seen by the detector. This is possible with daily budget-based recall metrics, by assuming that investigators review all user activity that day, even if the alert was generated by a single type of event. For example, an anomaly alert triggered by an unusual logon event might lead to an investigation which will uncover malicious email activity from the same user on the same day. This setting leads to metrics aligned with \cite{Tuor2017}. However keep in mind that the performance comparison is unfair since ADSAGE and other baselines see a single log data source.

ADSAGE and seq2one allow a flexible selection of features. However as our goal is to reduce feature engineering and preprocessing, we consistently use following approach for all events from CERT. First, we extract two time features from the date/time of each event: the minute of day and day of week. We represent both through their cosine and sine values in order to model their periodical nature. Second, we use all other (i.e. non time) available event attributes as is (with one hot encoding for categorical values). In each experiment, we list these additional features for completeness.

We also detail results for individual threat scenarios (section \ref{eval-scenarios}). This helps understanding which types of malicious behaviors are well detected by each method, and whether "blind spots" remain. Certain scenarios are virtually impossible to detect using some log data sources. For example, scenario 2 does not involve any logon activity, meaning that logon event detectors only cannot possibly alert about threats of this type.

For these reasons, methods presented in the following experiments should not be viewed as standalone, "one-fits-all" detectors. They rather are complementary, and each one addresses the CERT insider threat detection problem from a different perspective, depending on its data source. Though we compare our results to those of \cite{Tuor2017} (who performs detection at user-day level and uses all data sources), our focus is on understanding which event types are relevant (in general and for each scenario) and finding out whether insider threat detection is feasible at fine-grained event level.

\subsubsection{Detecting threats in logon events}\label{eval-cert-logon}
In a first experiment, we apply ADSAGE and other baselines to detect insider threats in logon events from the CERT dataset. In addition to edge sources and destinations and time features, we include a binary attribute indicating whether the action performed was a login or a logoff.

We use two simple rule-based baseline detectors. "Own PC" flags all logon events occurring on user's own machine (defined as the most used computer for this user) as normal; all other events are considered anomalous. "Known PC" considers a logon event to be normal if the corresponding user-computer edge was observed in the training set; otherwise it will be flagged as anomalous. Both methods provide binary decisions.

We use following hyperparameters for seq2one and ADSAGE's RNN: 1 layer of 30 LSTM units, 15 timesteps, batch size = 100, learning rate = 0.001 with decay factor of 0.5 after 1 epoch without improvement and the dimensionality of computer embeddings is set to 20. For ADSAGE's FFNN we use 3 layers of respectively 50, 30 and 10 units with relu activation and dropout set to 0.2. We perform 5 runs with 10 epochs.

\begin{figure}
	\centering
	\includegraphics[width=\columnwidth]{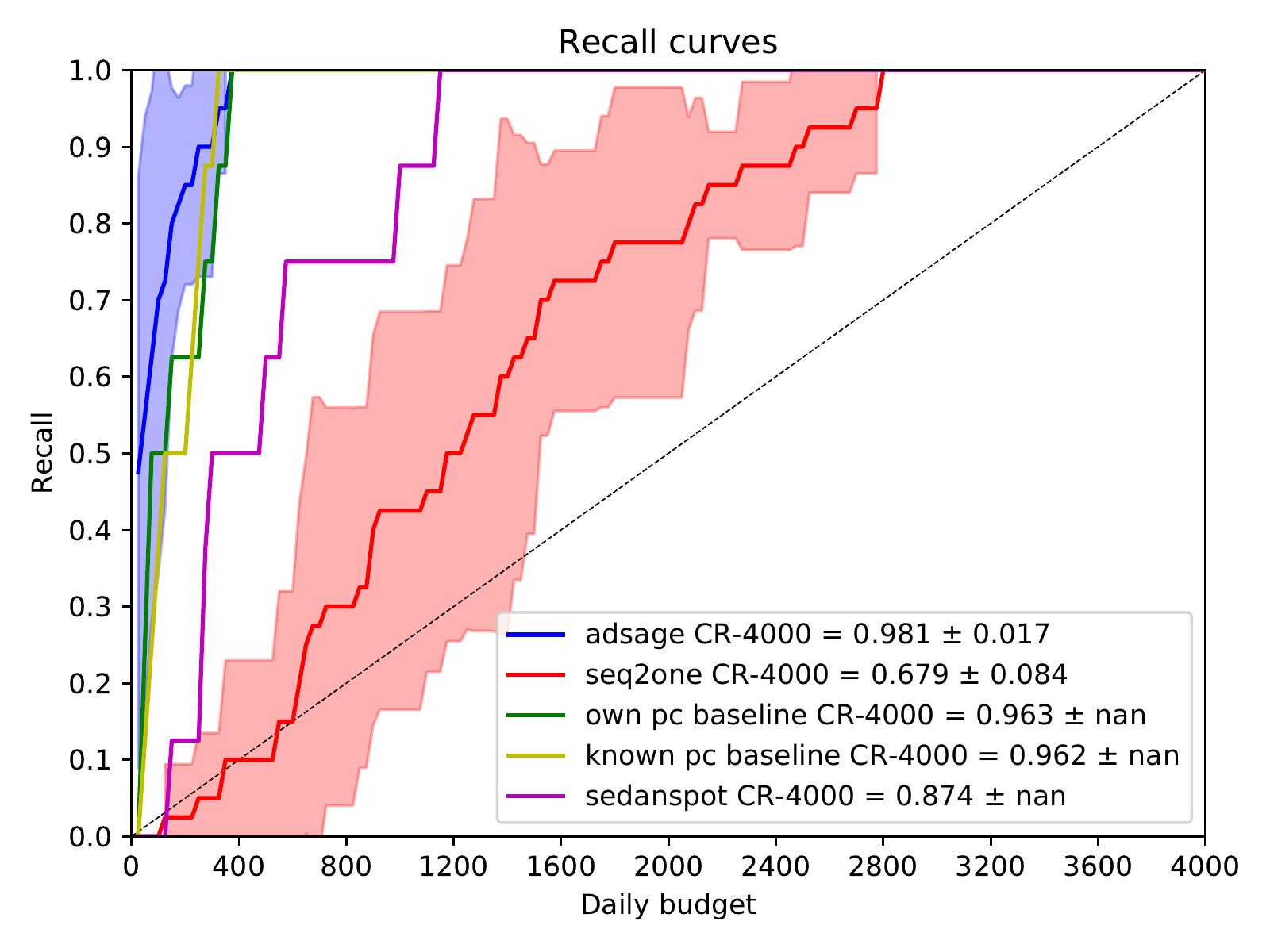}
	\caption{Recall curves with 95\% confidence intervals over 5 runs for detecting threats present in logon events.}
	\Description{The figure shows recall curves for detectors applied on logon events. ADSAGE, own pc and known pc have highest recall values, reaching recall = 1 for a daily budget of around 400.}
	\label{fig-logon-recall-curve}
\end{figure}

Detection results are shown in table \ref{tbl-results-logon}. When it comes to detecting threats present in logon events only, ADSAGE outperforms all other methods, with cumulative recall at maximum budget of 0.981. Cumulative recalls at lower budgets show a similar picture, and full recall curves presented in figure \ref{fig-logon-recall-curve} confirm that ADSAGE performs best at almost any budget. However, for the task of detecting all threats (i.e. including the ones not present in logon activity), ADSAGE is outperformed by the system of \cite{Tuor2017}.

\begin{table}[]
	\centering
	\caption{Detection results on logon events. For seq2one and ADSAGE we report 95\% confidence intervals over 10 runs. Top table: detecting threats present in logon events only, bottom table: detecting all threats (including those not present in logon events).}
	\label{tbl-results-logon}
	\begin{tabular}{@{}llll@{}}
		\toprule
		\textbf{Logon threats} & CR-400            & CR-1000           & CR-4000           \\ \midrule
		own pc                 & 0.633             & 0.853             & 0.963             \\
		known pc               & 0.617             & 0.847             & 0.962             \\
		SedanSpot              & 0.219             & 0.513             & 0.874             \\
		seq2one                & 0.039 $\pm$ 0.061 & 0.171 $\pm$ 0.151 & 0.679 $\pm$ 0.084 \\
		ADSAGE & \textbf{0.813 $\pm$ 0.172} & \textbf{0.925 $\pm$ 0.069} & \textbf{0.981 $\pm$ 0.017} \\ \midrule
		\textbf{All threats}   & CR-400            & CR-1000           & CR-4000           \\ \midrule
		own pc                 & 0.268             & 0.420             & \textbf{0.772}    \\
		known pc               & 0.280             & 0.426             & \textbf{0.772}    \\
		SedanSpot              & 0.119             & 0.338             & \textbf{0.814}    \\
		seq2one                & 0.047 $\pm$ 0.088 & 0.155 $\pm$ 0.073 & 0.679 $\pm$ 0.084 \\
		\cite{Tuor2017}        & \textbf{0.731}    & \textbf{0.893}    & not reported      \\
		ADSAGE & 0.432 $\pm$ 0.037          & 0.605 $\pm$ 0.102          & \textbf{0.842 $\pm$ 0.104} \\ \bottomrule
	\end{tabular}
\end{table}

\subsubsection{Detecting threats in email events}\label{eval-cert-email}
In a second experiment, we use email events as log data source. Email events from the CERT dataset represent an email being sent or received/read. Considering the significant overlap between the two, we only use "send" events. 

In addition to time features, we use following attributes from email events: email size (numeric), sender and receiver fields represented as embeddings ("from", "to", "cc", "bcc") and email content (text). Representing the sender is straightforward as it contains only one email address, so we use a simple embedding layer. However, receiver fields can contain several entities, so we combine them with an embedding bag layer \cite{embedding-bag} to obtain a fixed length representation. All three receiver fields are encoded as separate features; senders and receivers are embedded into a unique vector space. Text content of emails is represented through pre-trained word vectors, combined with a pooling scheme \cite{Wieting2015}. We have empirically determined that GloVe \cite{glove} vectors with average pooling work best for our problem.

We use two rule-based baselines for anomaly detection in email events. In the first, called "known receivers", each email event is assigned a score representing the proportion of unobserved receivers, i.e. receivers that were never contacted by the sender during the training period. The second is referred to as "known receiver set". It assigns a binary score depending on whether the exact set of receivers was observed in the training set for the corresponding sender (normal) or not (anomalous).

We use following hyperparameters for seq2one and ADSAGE's RNN: 1 layer of 100 LSTM units, 20 timesteps, batch size 1024, 5 epochs, learning rate of 0.01 with 0.5 decay factor after 1 epoch without improvement. Embeddings of email senders and receivers are of dimension 20. For ADSAGE's FFNN we use 3 layers of respectively 50, 30 and 10 units with relu activation and dropout = 0.2. We perform 5 runs of with 5 epochs and use the same data split as for logon events, but we train only on a random sample of all users (10\%). This speeds up the training process without significantly altering performance.

\begin{figure}
	\centering
	\includegraphics[width=\columnwidth]{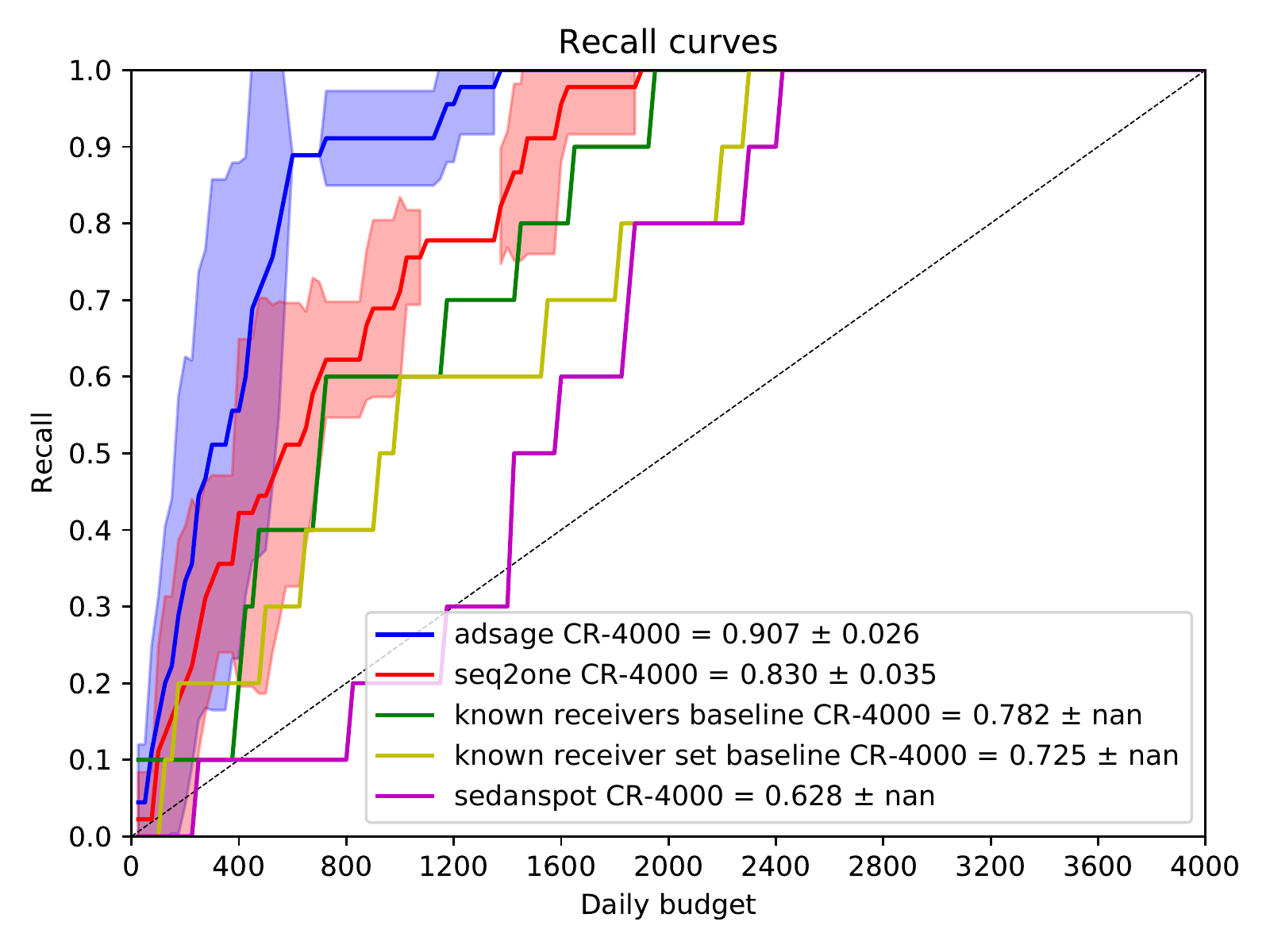}
	\caption{Recall curves with 95\% confidence intervals over 5 runs for detecting threats present in email events.}
	\Description{The figure shows recall curves for detectors applied on email events. ADSAGE has highest recall values, and reaches recall = 1 for a daily budget of around 1400.}
	\label{fig-email-recall-curve}
\end{figure}

Detection results are shown in table \ref{tbl-results-email}. For threats present in email events, ADSAGE outperforms other methods and reaches a cumulative recall at maximum budget CR-4000 = 0.907. As shown in figure \ref{fig-email-recall-curve}, a budget of around 800 allows to detect 90\% of threats in email events. Applying ADSAGE to email events also allows to detect threats present in all events effectively (CR-4000 = 0.930), even though the system of \cite{Tuor2017} still performs best. Nevertheless it suggests that email events are a good marker for insider threats.

\begin{table}[]
	\centering
	\caption{Detection results on email events. For seq2one and ADSAGE we report 95\% confidence intervals over 5 runs. Top table: scores when detecting threats present in email events only, bottom table: scores when detecting all threats (including those not present in email events).}
	\label{tbl-results-email}
	\begin{tabular}{@{}llll@{}}
		\toprule
		\textbf{Email threats} & CR-400            & CR-1000           & CR-4000           \\ \midrule
		known receivers        & \textbf{0.106}    & 0.340             & 0.782             \\
		known receiver set     & \textbf{0.138}    & 0.278             & 0.725             \\
		SedanSpot              & 0.044             & 0.098             & 0.628             \\
		seq2one & \textbf{0.217 $\pm$ 0.124} & 0.431 $\pm$ 0.109          & 0.830 $\pm$ 0.035          \\
		ADSAGE  & \textbf{0.332 $\pm$ 0.226} & \textbf{0.646 $\pm$ 0.117} & \textbf{0.907 $\pm$ 0.026} \\ \midrule
		\textbf{All threats}   & CR-400            & CR-1000           & CR-4000           \\ \midrule
		known receivers        & 0.116             & 0.408             & 0.827             \\
		known receiver set     & 0.134             & 0.318             & 0.754             \\
		SedanSpot              & 0.280             & 0.415             & 0.784             \\
		seq2one                & 0.199 $\pm$ 0.093 & 0.426 $\pm$ 0.100 & 0.822 $\pm$ 0.036 \\
		\cite{Tuor2017}        & \textbf{0.731}    & \textbf{0.893}    & not reported      \\
		ADSAGE  & 0.447 $\pm$ 0.118          & 0.728 $\pm$ 0.67           & \textbf{0.930 $\pm$ 0.017} \\ \bottomrule
	\end{tabular}
\end{table}

\subsubsection{Detecting threats in web events}\label{eval-cert-http}
In a third experiment on the CERT dataset, we use web events as data source. Web events represent user browsing activities. In addition to edge sources and destinations and time features, we tried adding the content of web page as text feature but ended up discarding it because it did not improve detection performance significantly.

We use a rule-based baseline for anomaly detection which we call "known domain". It assigns binary anomaly scores based on whether the web domain of an event has been observed in the training period for the corresponding user. Thus all accesses to new, unobserved domains are considered anomalous; the rest is deemed normal.

We use following hyperparameters for seq2one and ADSAGE's RNN: 1 layer of 100 LSTM units, 20 timesteps, batch size 2048, 5 epochs and learning rate of 0.001 with 0.5 decay factor after 1 epoch without improvement. Embeddings of email senders and receivers are of dimension 50. For ADSAGE's FFNN we use 3 layers of respectively 50, 30 and 10 units with relu activation with dropout = 0.2. As the volume of web events is much larger than other for other audit data sources, we train on a random sample of 5\% of all users, discarding malicious ones and perform 5 runs with 5 epochs.

\begin{figure}
	\centering
	\includegraphics[width=\columnwidth]{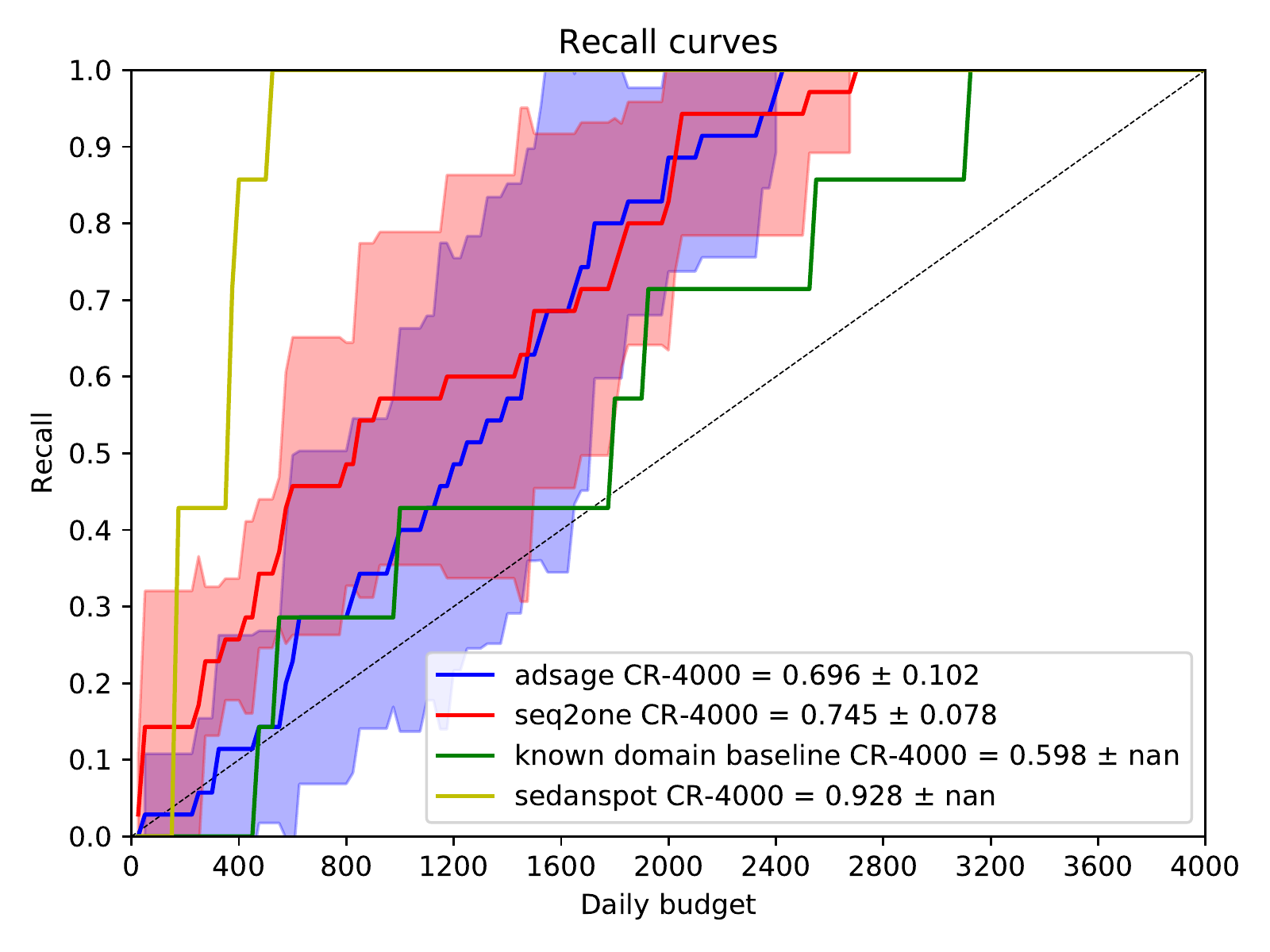}
	\caption{Recall curves with 95\% confidence intervals over 5 runs for detecting threats present in web events.}
	\Description{The figure shows recall curves for detectors applied on web events. SedanSpot has highest recall values, reaching recall = 1 for a daily budget of around 500.}
	\label{fig-web-recall-curve}
\end{figure}

Table \ref{tbl-results-web} shows detection results. SedanSpot outperforms other methods with CR-4000 = 0.928 when detecting threats present in web events only. As shown on figure \ref{fig-web-recall-curve}, a budget of around 600 is sufficient to detect all threats in web events. When considering recall of all threats, SedanSpot gives the best results (followed by seq2one, difference is not statistically significant), but is not as effective as the system from \cite{Tuor2017}, which uses all data sources. Overall, ADSAGE is not adapted to detect anomalies in web events represented as user to web domain edges. One possible explanation is that the domain identifier is not informative enough to characterize browsing behavior.

\begin{table}[]
	\centering
	\caption{Detection results on web events. For seq2one and ADSAGE we report 95\% confidence intervals over 5 runs. Top table: scores when detecting threats present in web events only, bottom table: scores when detecting all threats (including those not present in web events).}
	\label{tbl-results-web}
	\begin{tabular}{@{}llll@{}}
		\toprule
		\textbf{Web threats} & CR-400            & CR-1000           & CR-4000                    \\ \midrule
		known domain         & 0                 & 0.150             & 0.598                      \\
		SedanSpot            & \textbf{0.313}    & \textbf{0.711}    & \textbf{0.928}             \\
		seq2one              & 0.175 $\pm$ 0.129 & 0.344 $\pm$ 0.054 & 0.745 $\pm$ 0.078          \\
		ADSAGE               & 0.054 $\pm$ 0.070 & 0.179 $\pm$ 0.127 & 0.696 $\pm$ 0.102          \\ \midrule
		\textbf{All threats} & CR-400            & CR-1000           & CR-4000                    \\ \midrule
		known domain         & 0.042             & 0.148             & 0.588                      \\
		SedanSpot            & 0.199             & 0.432             & \textbf{0.736}             \\
		seq2one              & 0.132 $\pm$ 0.078 & 0.259 $\pm$ 0.087 & \textbf{0.693 $\pm$ 0.061} \\
		\cite{Tuor2017}      & \textbf{0.731}    & \textbf{0.893}    & not reported               \\
		ADSAGE               & 0.035 $\pm$ 0.030 & 0.109 $\pm$ 0.026 & 0.608 $\pm$ 0.031          \\ \bottomrule
	\end{tabular}
\end{table}

\subsubsection{Results by threat scenarios}\label{eval-scenarios}
In order to characterize which methods and data sources allow to detect each CERT insider threat scenario (see section \ref{CERT_dataset}), we evaluate all logon, email and web detectors using a different data split. We use the period from January to July 2010 as train set and test on August 2010 to April 2011. This allows us to assess detection performance on all threat scenarios, whereas the test set used by \cite{Tuor2017} contains only scenarios 2 and 4. Hyperparameter values determined earlier are kept unchanged.

Detection results (CR-4000 scores) for each scenario presented by detector and data source are shown in table \ref{tbl-results-all-scenarios}. It appears that monitoring logon events can be effective (CR-4000 $\geq$ 0.85) to detect scenarios 3, 4 and 5. Rule-based methods ("own PC", "known PC") give good performance for these scenarios, however ADSAGE and SedanSpot can be better for 3 and 5 respectively. Email traffic is a good audit data source to detect scenarios 2, 4 and 5. ADSAGE ranks among the best detectors for scenarios 2 to 5, while SedanSpot is particularly effective for scenario 2 and the "known receivers" baseline proves strong against scenarios 4 and 5. Finally, web browsing logs can be used to uncover scenarios 2 (using SedanSpot or "known domain"), 4 and 5 (with "known domain").

\begin{table*}[]
	\centering
	\caption{Detection performance (cumulative recall at maximum budget, CR-4000) for different insider threat scenarios.}
	\label{tbl-results-all-scenarios}
	\begin{tabular}{@{}llccccc@{}}
		\toprule
		&                             & \multicolumn{5}{c}{Threat scenario}                                                                                                            \\
		Data source                                                                 & Detection method            & 1                          & 2                          & 3                          & 4                          & 5                          \\ \midrule
		\multirow{5}{*}{\begin{tabular}[c]{@{}l@{}}Logon \\ (10 runs)\end{tabular}} & own pc baseline             & 0.392                      & \textbf{0.636}             & 0.848                      & 0.966                      & 0.963                      \\
		& known pc baseline           & \textbf{0.598}             & \textbf{0.646}             & \textbf{0.855}             & 0.963                      & 0.963                      \\
		& SedanSpot                   & 0.117                      & \textbf{0.642}             & 0.364                      & 0.875                      & \textbf{0.969}             \\
		& seq2one                     & \textbf{0.618 $\pm$ 0.099} & \textbf{0.614 $\pm$ 0.019} & 0.695 $\pm$ 0.017          & 0.690 $\pm$ 0.046          & 0.515 $\pm$ 0.185          \\
		& ADSAGE                      & \textbf{0.645 $\pm$ 0.109} & \textbf{0.667 $\pm$ 0.086} & 0.825 $\pm$ 0.012          & \textbf{0.975 $\pm$ 0.007} & 0.495 $\pm$ 0.135          \\ \midrule
		\multirow{5}{*}{\begin{tabular}[c]{@{}l@{}}Email \\ (5 runs)\end{tabular}}  & known receivers baseline    & 0.652                      & 0.810                      & 0.600                      & \textbf{0.885}             & \textbf{0.900}             \\
		& known receiver set baseline & 0.617                      & 0.714                      & 0.646                      & 0.783                      & 0.894                      \\
		& SedanSpot                   & \textbf{0.823}             & \textbf{0.928}             & \textbf{0.664}             & 0.548                      & 0.763                      \\
		& seq2one                     & \textbf{0.762 $\pm$ 0.124} & 0.770 $\pm$ 0.036          & \textbf{0.669 $\pm$ 0.054} & \textbf{0.815 $\pm$ 0.032} & 0.730 $\pm$ 0.210          \\
		& ADSAGE                      & 0.668 $\pm$ 0.120          & \textbf{0.853 $\pm$ 0.132} & \textbf{0.669 $\pm$ 0.043} & \textbf{0.798 $\pm$ 0.138} & \textbf{0.942 $\pm$ 0.071} \\ \midrule
		\multirow{4}{*}{\begin{tabular}[c]{@{}l@{}}Web\\ (5 runs)\end{tabular}}     & known domain baseline       & \textbf{0.731}             & 0.853                      & \textbf{0.717}             & \textbf{0.837}             & \textbf{1.000}             \\
		& SedanSpot                   & 0.294                      & \textbf{0.944}             & 0.554                      & 0.433                      & 0.906                      \\
		& seq2one                     & \textbf{0.743 $\pm$ 0.089} & 0.720 $\pm$ 0.021          & 0.619 $\pm$ 0.029          & 0.638 $\pm$ 0.039          & 0.680 $\pm$ 0.257          \\
		& ADSAGE                      & 0.468 $\pm$ 0.075          & 0.704 $\pm$ 0.112          & \textbf{0.727 $\pm$ 0.029} & 0.446 $\pm$ 0.056          & 0.631 $\pm$ 0.287          \\ \bottomrule
	\end{tabular}
\end{table*}

These results suggest that anomalies flagged by distinct detectors overlap only partially, thus methods can be complementary in detecting insider threats. By combining anomaly scores obtained from several perspectives (i.e. computed by distinct methods using different audit data sources), we can expect detection performance improvement. Possible approaches to perform fine-grained anomaly score fusion are mentioned in section \ref{extensions}. We insist on the fact that this approach differs from data aggregation: as anomaly scores are attributed at fine-grained level, the root cause of alerts can still be determined precisely.

\subsection{Detecting anomalies in real authentications}\label{eval-lanl}
To complement our results on the synthetic CERT datasets, we evaluate our methods on real-world authentication logs from the LANL's multi-source cybersecurity events \cite{LANL-dataset, LANL-paper}. This dataset contains Windows authentications, process traces, DNS data and network flows of from more than 12000 users collected over 58 days. However we only use authentication events, as they contain ground truth anomalies (malicious examples injected by a red team) unlike other traces.

We preprocess the dataset as follows. First, we remove authentications from special aliases and system accounts. Second, we align attributes of normal and red team events by keeping only the timestamp, user, source and destination computer attributes. For normal authentications events we use the "source" user attribute, ignoring the "destination" user. This is justified because the values of these two fields are the same most of the time, except if source user A authenticates as destination user B. In this case, A will be seen as B after such authentication. Finally, we merge normal and red team events to obtain a dataset containing almost 12000 users. Our maximal budget for cumulative recall metrics is therefore set at 12000. We use days 1 to 8 as train set (44.2M events, 50 anomalies) and days 9 to 13 (27.7M events, 587 anomalies) as test set.

For ADSAGE and seq2one, we use the same time features as for CERT data. Graph features are the source and destination computer of an authentication event, meaning that we have two attributed user to computer edges. For this reason, we implement two rule-based baselines "known source PC" and "known destination PC", which are equivalent to "known PC" for logon events (see section \ref{eval-cert-logon}) for each corresponding graph feature. We also run two instances of SedanSpot, one for edges from user to source computer and the other for user to destination computer.

We use following hyperparameters for seq2one and ADSAGE's RNN: 1 layer of 50 LSTM units, 10 timesteps, batch size 512, 15 epochs, learning rate of 0.001 with 0.5 decay factor after 1 epoch without improvement and no dropout. Embeddings of source and destination computers have a dimensionality of 20. For ADSAGE's FFNN we use 3 layers of respectively 50, 30 and 10 units with relu activation with dropout = 0.2. We train on a 10\% random sample of all users.

Cumulative recall values at budgets 1000, 4000 and 12000 are presented in table \ref{tbl-results-lanl}. ADSAGE and the "known source pc" rule-based classifier outperform all other methods at all 3 budget values. At maximum budget, their cumulative recall reaches 0.88 and 0.89 respectively (though the difference is not statistically significant). Detection results from SedanSpot and our rule-based classifier also suggest that the source computer attribute in LANL authentication events is more informative than the destination computer. ADSAGE has the advantage to support both attributes simultaneously. Note that how results obtained on LANL and CERT logon events are consistent, which is reassuring given that unlike LANL authentications, the CERT datasets are synthetic.

\begin{table}[]
	\caption{Detection results for red team anomalies in LANL authentication events. We report normalized cumulative recalls at budgets 1000, 4000 and 12000 for 5.2. For seq2one and ADSAGE we report 95\% confidence intervals over 5 runs.}
	\label{tbl-results-lanl}
	\begin{tabular}{llll}
		\hline
		\begin{tabular}[c]{@{}l@{}}Detecting red team events\\ in LANL authentications\end{tabular} &
		CR-1000 &
		CR-4000 &
		CR-12000 \\ \hline
		known dest pc         & 0.237          & 0.566          & 0.829          \\
		known source pc       & \textbf{0.254} & \textbf{0.669} & \textbf{0.890} \\
		SedanSpot (dest) pc   & 0.016          & 0.104          & 0.490          \\
		SedanSpot (source) pc & 0.089          & 0.191          & 0.538          \\
		seq2one &
		\begin{tabular}[c]{@{}l@{}}0.167 \\ $\pm$ 0.023\end{tabular} &
		\begin{tabular}[c]{@{}l@{}}0.423 \\ $\pm$ 0.031\end{tabular} &
		\begin{tabular}[c]{@{}l@{}}0.752 \\ $\pm$ 0.014\end{tabular} \\
		ADSAGE &
		\textbf{\begin{tabular}[c]{@{}l@{}}0.255 \\ $\pm$ 0.078\end{tabular}} &
		\textbf{\begin{tabular}[c]{@{}l@{}}0.653 \\ $\pm$ 0.042\end{tabular}} &
		\textbf{\begin{tabular}[c]{@{}l@{}}0.877 \\ $\pm$ 0.014\end{tabular}} \\ \hline
	\end{tabular}
\end{table}

\section{Related work}
\subsection{CERT insider threat use case}\label{SOTA_CERT}
Despite a large body of work addressing the CERT use case, comparing detection performance of existing insider threat detection systems remains challenging, due to different metrics and choices of train/test data split. In most cases, ROC AUC score (as indicator of detection performance across all decision thresholds) or detection and false positive rate (for a single decision threshold) is used. Highest ROC AUC is reported by \cite{Hall2018} (0.99) however they use a very limited data subset and rely on information unavailable in practice (Does user quit at later time?), so whether this level of performance could be attained on the whole dataset remains an open question. \citet{Yuan2018a} report a ROC AUC of 0.95 using features designed with expert knowledge about threats (see table 1 in their paper). When considering detection and false positive rate, the best score is reported by \cite{Le2018b}, achieving 86\% detected threats with 20\% false positives (and 0.86 ROC AUC). However because of the very high class imbalance, the false positive rate is much too high in practice (e.g. 1 million samples would generate 200,000 false alarms). To circumvent the absence of standard benchmark, we adopt the evaluation setting of \citet{Tuor2017}, which relies on business-realistic recall metrics (see section \ref{metrics} for more details).

\subsection{Graph and text for insider threat detection}
Graph features have proven useful in insider threat detection, for example to represent email activity \cite{Eberle2010, Okolica2008}, collaboration through access logs \cite{Chen2012} or workgroup roles \cite{Nance2011}. Graph analysis can be used to address concept drift \cite{Parveen2011} and community detection \cite{Senator2013}. Additionally, provenance and knowledge graphs can prevent insider threats from a physical security perspective \cite{Mavroeidis2018, Althebyan2007, Nwafor2018}. Text features can help characterizing user sentiment \cite{Kandias2013, Brown2013, Mayhew2015}.

Surprisingly, in the CERT datasets \cite{Glasser2013, CERT_dataset} graph (user to computer relations, web pages, email communication, LDAP attributes) and text features (content of web pages, emails and files) have been largely ignored, except for computer relations in logon events. To the best of our knowledge, the only system making use of the web page graph is \cite{Gamachchi2015}, which clusters users and web pages to find similarities in browsing behavior. Unfortunately, no detection performance is reported. For email traffic, existing systems use addresses to determine if receivers are internal or external \cite{Agrafiotis2014, Rashid2016, Le2018a}, but do not attempt to use the full email graph. Graph features from the LDAP attributes are more often used \cite{Tuor2017, Legg2015, Lv2018, Le2018a, Le2018b, Hall2018}. Text content is only used in \cite{Gavai2015} as simple statistical features and in \cite{Legg2015} through bag-of-words and linguistic features.

\subsection{Graph edge level anomaly detection}
\label{SOTA_anomaly_detection}

Like ADSAGE, some methods perform anomaly detection at graph edge level, but we are not aware of another method supporting both sequences of edges and edge attributes. Existing works perform anomaly detection in edge streams, but do not support edge attributes \cite{Eswaran2018, Yu2018, Zheng2019, Heard2010, Ranshous2016, Yoon2019} . On the contrary, EdgeCentric \cite{Shah2016} supports edge attributes, however it detects anomalies at node level. Unfortunately not all methods were compared to each other. In the end, we have chosen to use SedanSpot \cite{Eswaran2018} as baseline for its good performance and usable, well-documented implementation.

\subsection{Anomaly detection at event level}
A main characteristic of our approach is to perform detection at fine-grained event level to enhance alert traceability. If this perspective is novel for insider threats, similar works can be found in system log analysis. DeepLog \cite{Du2017} uses workflows of normal behavior to diagnose which element is anomalous within log line sequences while \citet{Wurzenberger2017} use clustering for similar purposes. \citet{Brown2018} go further and provide character level anomaly scores by using neural network attention mechanisms.

\section{Conclusion}
\subsection{Main findings}
ADSAGE fills a gap in anomaly detection at graph edge level, as existing methods do not support sequences and attributed edges simultaneously. Our method supports heterogeneous attributes: numeric, categorical and text. Focusing on insider threat detection, we have benchmarked ADSAGE against SedanSpot and other baselines on different data sources from the CERT and LANL datasets (authentication, email and web browsing logs). Our approach significantly differs from concurrent systems in that detection is performed at fine-grained event to enhance alert traceability. We have found that ADSAGE is effective to detect anomalies in authentications (represented as user to computer edges) and in email traffic (sender to receiver edges). For CERT web browsing logs, represented as user to web domain relations, ADSAGE is not appropriate but other methods such as SedanSpot or rule-based detectors can be used instead. Crucially, we note that results obtained on authentication logs are consistent across LANL (real) and CERT (synthetic) datasets, which is reassuring concerning realism of the latter.

As our method uses only one audit data source at a time, reporting results split by threat scenarios has allowed us to gain insight about which audit data sources and which methods are suited to target specific malicious behaviors. Although we could not meet state-of-the-art performance of \cite{Tuor2017} when detecting threats over all audit source domains, a direct comparison is unfair as our method relies on a unique audit data source. Still, we believe the performance gap is encouraging given that preprocessing and feature engineering effort are reduced, while alert traceability is improved.

Overall, our experimental results show that insider threat detection at fine-grained event level is feasible. Beyond the choice of detection method, we have found that graph (user to computer relations, email communications) and text features (email contents) from the CERT datasets can be informative to spot insider threats.

\subsection{Possible extensions}\label{extensions}
Concerning ADSAGE specifically, we have chosen to generate one negative sample for each positive one. We suggest to conduct further experiments to assess the influence of negative sampling rate. 

More generally, this work performs insider threat detection using only one data source at a time. Detection results by threat scenarios show that anomalies retrieved by different methods overlap only partially, suggesting that detection improvement can be expected from  combining several detectors. In this regard, detailed performance results split by threat scenario presented here could help choosing complementary methods to be combined. One simplistic possibility is to aggregate anomaly scores, for instance with averaging at user-day level. A second, more sophisticated approach, could be to run several synchronized instances of ADSAGE (one for each audit data source) sharing their RNN states. This could help provide context from other data sources while still performing fine-grained anomaly detection.

%

%
\bibliographystyle{ACM-Reference-Format}
\bibliography{adsage}

%

\end{document}